\documentclass{article}
\usepackage{ijcai18}

\usepackage{times}
\usepackage{xcolor}
\usepackage{soul}
\usepackage[utf8]{inputenc}
\usepackage[small]{caption}

\usepackage{graphics,epsfig,amsmath,amsfonts,amssymb,times,colortbl,subfigure,floatflt,wrapfig,enumerate,tikz,url,xcolor}
\usepackage{multirow}
\usepackage{algorithmic,eqparbox,comment}
\usepackage{bm}
\usepackage{booktabs}
\usepackage[symbol]{footmisc}
\usepackage{algorithm, algorithmic}
\usepackage{array}

\newtheorem{definition}{Definition}

\newcommand{\mat}[1]{\bm{#1}}
\newcommand{\ten}[1]{\bm{\mathcal{#1}}}

\title{A Support Tensor Train Machine}

\author{
Cong Chen$^1$, 
Kim Batselier$^1$, 
Ching-Yun Ko$^1$, 
Ngai Wong$^1$, 
\\ 
$^1$ The Department of Electrical and Electronic Engineering, The University of Hong Kong \\
chencong@eee.hku.hk,
kimb@eee.hku.hk,
cyko@eee.hku.hk,
nwong@eee.hku.hk
}
\begin{document}
\maketitle
\begin{abstract}
There has been growing interest in extending traditional vector-based machine learning techniques to their tensor forms. An example is the support tensor machine (STM) that utilizes a rank-one tensor to capture the data structure, thereby alleviating the overfitting and curse of dimensionality problems in the conventional support vector machine (SVM). However, the expressive power of a rank-one tensor is restrictive for many real-world data. To overcome this limitation, we introduce a support tensor train machine (STTM) by replacing the rank-one tensor in an STM with a tensor train. Experiments validate and confirm the superiority of an STTM over the SVM and STM.  
\end{abstract}

\section{Introduction}
Classification algorithm design has been a popular topic in machine learning, pattern recognition and computer vision for decades. One of the most representative and successful classification algorithms is the support vector machines (SVM)~\cite{vapnik2013nature}, which achieves an enormous success in pattern classification by minimizing the Vapnik-Chervonenkis dimensions and structural risk. However, a standard SVM model is based on vector inputs and cannot directly deal with matrices or higher dimensional data structures, namely, tensors, which are very common in real-life applications. For example, a grayscale picture is stored as a matrix which is a second-order tensor, while color pictures have a color axis and are naturally third-order tensors. The common SVM realization on such high dimensional inputs is by reshaping each sample into a vector. 
 However, when the training data sample size is relatively small compared to the feature vector dimension, it may easily result in poor classification performance due to overfitting~\cite{li2006multitraining,tao2006asymmetric,yan2007multilinear}. To overcome this, researchers have focused on exploring new data structures and corresponding numerical operations. A versatile data structure is tensors, which have recently received much attention in the machine learning community. In particular, tensor trains have found various applications. In~\cite{chen2017parallelized} a tensor train based polynomial classifier is proposed that encodes the coefficients of the polynomial as a tensor train. In~\cite{novikov2015tensorizing} tensor trains are used to compress the traditional fully connected layers of a neural network into fewer number of parameters. Tensor trains have also been used to represent nonlinear predictors~\cite{novikov2016exponential} and classifiers~\cite{miles2016supervised}. Moreover, the canonical polyadic (CP) tensor decomposition has been used for speeding up the convolution step in convolutional neural networks~\cite{lebedev2014speeding} and the Tucker decomposition for the classification of tensor data~\cite{signoretto2014learning} etc.   
 
Not surprisingly, standard SVMs have also been extended to tensor formulations yielding significant performance enhancements~\cite{tao2005supervised,kotsia2011support}. Reference~\cite{tao2005supervised} proposes a supervised tensor learning (STL) scheme by replacing the vector inputs with tensor inputs and decomposing the corresponding weight vector into a rank-1 tensor, which is trained by the alternating projection optimization method. Based on this learning scheme,~\cite{Tao2007} extends the standard linear SVM to a general tensor form called the support tensor machine (STM). Although STM lifts the overfitting problem in traditional SVMs, the expressive power of a rank-1 weight tensor is low, which translates into an often poor classification accuracy. In~\cite{kotsia2011support} and ~\cite{kotsia2012higher}, the rank-1 weight tensor of STM is generalized to Tucker and CP forms for stronger model expressive power. However, the determination of a good CP-rank is NP-complete~\cite{Hastad1990} and the number of parameters in the Tucker form is exponentially large, which still suffers from the curse of dimensionality. 

This work proposes, for the first time, a support tensor train machine (STTM) wherein the rank-1 weight tensor of an STM is replaced by a tensor train that can approximate any tensor with a scalable number of parameters. An STTM has the following advantages: 
\begin{enumerate}
\item With a small sample size, STTM has comparable or better classification accuracy than the standard SVM.
\item The expressive power of a tensor train increases with its tensor train ranks. This means an STTM can capture much richer structural information than an STM and lead to improved classification accuracy.
\item A tensor train mixed-canonical form can be exploited to further speed up algorithmic convergence. 
\end{enumerate}
In the following, Section~\ref{sec:prelim} introduces the necessary tensor basics and the key ideas of the SVM and STM frameworks. The proposed STTM is presented in Section~\ref{sec:STTM}. Experiments are given in Section~\ref{sec:experiments} to show the advantages of an STTM over SVM and STM. Finally, Section~\ref{sec:conclusion} draws the conclusions.



\section{Preliminaries}
\label{sec:prelim}
\subsection{Tensor Basics}
\begin{figure}[tb] 
\begin{center} 
\ifx\du\undefined
  \newlength{\du}
\fi
\setlength{\du}{15\unitlength}
\begin{tikzpicture}
\pgftransformxscale{0.750000}
\pgftransformyscale{-0.700000}
\definecolor{dialinecolor}{rgb}{0.000000, 0.000000, 0.000000}
\pgfsetstrokecolor{dialinecolor}
\definecolor{dialinecolor}{rgb}{1.000000, 1.000000, 1.000000}
\pgfsetfillcolor{dialinecolor}
\pgfsetlinewidth{0.00000\du}
\pgfsetdash{}{0pt}
\pgfsetdash{}{0pt}
\pgfsetmiterjoin
\definecolor{dialinecolor}{rgb}{0.000000, 0.000000, 0.000000}
\pgfsetstrokecolor{dialinecolor}
\pgfpathellipse{\pgfpoint{13.375000\du}{11.975000\du}}{\pgfpoint{1.575000\du}{0\du}}{\pgfpoint{0\du}{1.525000\du}}
\pgfusepath{stroke}
\definecolor{dialinecolor}{rgb}{0.000000, 0.000000, 0.000000}
\pgfsetstrokecolor{dialinecolor}
\node at (13.375000\du,12.215000\du){};
\pgfsetlinewidth{0.00000\du}
\pgfsetdash{}{0pt}
\pgfsetdash{}{0pt}
\pgfsetbuttcap
{
\definecolor{dialinecolor}{rgb}{0.000000, 0.000000, 0.000000}
\pgfsetfillcolor{dialinecolor}
\definecolor{dialinecolor}{rgb}{0.000000, 0.000000, 0.000000}
\pgfsetstrokecolor{dialinecolor}
\draw (13.375000\du,13.500000\du)--(13.350000\du,15.950000\du);
}
\pgfsetlinewidth{0.00000\du}
\pgfsetdash{}{0pt}
\pgfsetdash{}{0pt}
\pgfsetmiterjoin
\definecolor{dialinecolor}{rgb}{0.000000, 0.000000, 0.000000}
\pgfsetstrokecolor{dialinecolor}
\pgfpathellipse{\pgfpoint{8.930000\du}{11.975000\du}}{\pgfpoint{1.575000\du}{0\du}}{\pgfpoint{0\du}{1.525000\du}}
\pgfusepath{stroke}
\definecolor{dialinecolor}{rgb}{0.000000, 0.000000, 0.000000}
\pgfsetstrokecolor{dialinecolor}
\node at (8.930000\du,12.215000\du){};
\pgfsetlinewidth{0.00000\du}
\pgfsetdash{}{0pt}
\pgfsetdash{}{0pt}
\pgfsetmiterjoin
\definecolor{dialinecolor}{rgb}{0.000000, 0.000000, 0.000000}
\pgfsetstrokecolor{dialinecolor}
\pgfpathellipse{\pgfpoint{17.780000\du}{12.000000\du}}{\pgfpoint{1.575000\du}{0\du}}{\pgfpoint{0\du}{1.500000\du}}
\pgfusepath{stroke}
\definecolor{dialinecolor}{rgb}{0.000000, 0.000000, 0.000000}
\pgfsetstrokecolor{dialinecolor}
\node at (17.780000\du,12.240000\du){};
\pgfsetlinewidth{0.00000\du}
\pgfsetdash{}{0pt}
\pgfsetdash{}{0pt}
\pgfsetbuttcap
{
\definecolor{dialinecolor}{rgb}{0.000000, 0.000000, 0.000000}
\pgfsetfillcolor{dialinecolor}
\definecolor{dialinecolor}{rgb}{0.000000, 0.000000, 0.000000}
\pgfsetstrokecolor{dialinecolor}
\draw (18.893693\du,13.060660\du)--(18.893693\du,16.000000\du);
}
\pgfsetlinewidth{0.00000\du}
\pgfsetdash{}{0pt}
\pgfsetdash{}{0pt}
\pgfsetbuttcap
{
\definecolor{dialinecolor}{rgb}{0.000000, 0.000000, 0.000000}
\pgfsetfillcolor{dialinecolor}
\definecolor{dialinecolor}{rgb}{0.000000, 0.000000, 0.000000}
\pgfsetstrokecolor{dialinecolor}
\draw (16.666307\du,13.060660\du)--(16.666307\du,16.00000\du);
}
\pgfsetlinewidth{0.00000\du}
\pgfsetdash{}{0pt}
\pgfsetdash{}{0pt}
\pgfsetmiterjoin
\definecolor{dialinecolor}{rgb}{0.000000, 0.000000, 0.000000}
\pgfsetstrokecolor{dialinecolor}
\pgfpathellipse{\pgfpoint{22.080000\du}{11.975000\du}}{\pgfpoint{1.575000\du}{0\du}}{\pgfpoint{0\du}{1.525000\du}}
\pgfusepath{stroke}
\definecolor{dialinecolor}{rgb}{0.000000, 0.000000, 0.000000}
\pgfsetstrokecolor{dialinecolor}
\node at (22.080000\du,12.215000\du){};
\pgfsetlinewidth{0.00000\du}
\pgfsetdash{}{0pt}
\pgfsetdash{}{0pt}
\pgfsetbuttcap
{
\definecolor{dialinecolor}{rgb}{0.000000, 0.000000, 0.000000}
\pgfsetfillcolor{dialinecolor}
\definecolor{dialinecolor}{rgb}{0.000000, 0.000000, 0.000000}
\pgfsetstrokecolor{dialinecolor}
\draw (23.193693\du,13.053338\du)--(23.193693\du,16.000000\du);
}
\pgfsetlinewidth{0.00000\du}
\pgfsetdash{}{0pt}
\pgfsetdash{}{0pt}
\pgfsetbuttcap
{
\definecolor{dialinecolor}{rgb}{0.000000, 0.000000, 0.000000}
\pgfsetfillcolor{dialinecolor}
\definecolor{dialinecolor}{rgb}{0.000000, 0.000000, 0.000000}
\pgfsetstrokecolor{dialinecolor}
\draw (20.966307\du,13.053338\du)--(20.966307\du,16.000000\du);
}
\pgfsetlinewidth{0.00000\du}
\pgfsetdash{}{0pt}
\pgfsetdash{}{0pt}
\pgfsetbuttcap
{
\definecolor{dialinecolor}{rgb}{0.000000, 0.000000, 0.000000}
\pgfsetfillcolor{dialinecolor}
\definecolor{dialinecolor}{rgb}{0.000000, 0.000000, 0.000000}
\pgfsetstrokecolor{dialinecolor}
\draw (22.080000\du,13.500000\du)--(22.080000\du,16.000000\du);
}
\definecolor{dialinecolor}{rgb}{0.000000, 0.000000, 0.000000}
\pgfsetstrokecolor{dialinecolor}
\node[anchor=west] at (8.430000\du,11.975000\du){$a$};
\definecolor{dialinecolor}{rgb}{0.000000, 0.000000, 0.000000}
\pgfsetstrokecolor{dialinecolor}
\node[anchor=west] at (9.200000\du,11.900000\du){};
\definecolor{dialinecolor}{rgb}{0.000000, 0.000000, 0.000000}
\pgfsetstrokecolor{dialinecolor}
\node[anchor=west] at (13.005000\du,11.975000\du){$\mat{a}$};
\definecolor{dialinecolor}{rgb}{0.000000, 0.000000, 0.000000}
\pgfsetstrokecolor{dialinecolor}
\node[anchor=west] at (17.200000\du,12.000000\du){$\mat{A}$};
\definecolor{dialinecolor}{rgb}{0.000000, 0.000000, 0.000000}
\pgfsetstrokecolor{dialinecolor}
\node[anchor=west] at (21.450000\du,12.00000\du){$\ten{A}$};
\end{tikzpicture}
\caption{Graphical representation of a scalar $a$, vector $\mat{a}$, matrix $\mat{A}$, and third-order tensor $\ten{A}$.}
\label{fig:graphical}
\end{center}
\end{figure}
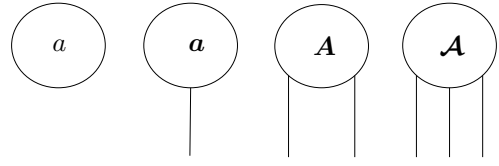

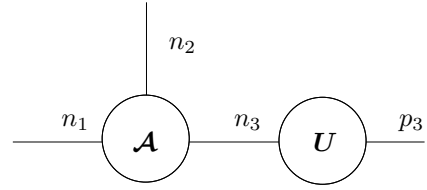
\begin{figure}[tb] 
\begin{center} 
\ifx\du\undefined
  \newlength{\du}
\fi
\setlength{\du}{15\unitlength}
\begin{tikzpicture}
\pgftransformxscale{0.750000}
\pgftransformyscale{-0.750000}
\definecolor{dialinecolor}{rgb}{0.000000, 0.000000, 0.000000}
\pgfsetstrokecolor{dialinecolor}
\definecolor{dialinecolor}{rgb}{1.000000, 1.000000, 1.000000}
\pgfsetfillcolor{dialinecolor}
\pgfsetlinewidth{0.00000\du}
\pgfsetdash{}{0pt}
\pgfsetdash{}{0pt}
\pgfsetbuttcap
\pgfsetmiterjoin
\pgfsetlinewidth{0.00000\du}
\pgfsetbuttcap
\pgfsetmiterjoin
\pgfsetdash{}{0pt}
\definecolor{dialinecolor}{rgb}{1.000000, 1.000000, 1.000000}
\pgfsetfillcolor{dialinecolor}
\pgfpathellipse{\pgfpoint{6.573437\du}{8.535937\du}}{\pgfpoint{1.464062\du}{0\du}}{\pgfpoint{0\du}{1.464062\du}}
\pgfusepath{fill}
\definecolor{dialinecolor}{rgb}{0.000000, 0.000000, 0.000000}
\pgfsetstrokecolor{dialinecolor}
\pgfpathellipse{\pgfpoint{6.573437\du}{8.535937\du}}{\pgfpoint{1.464062\du}{0\du}}{\pgfpoint{0\du}{1.464062\du}}
\pgfusepath{stroke}
\pgfsetbuttcap
\pgfsetmiterjoin
\pgfsetdash{}{0pt}
\definecolor{dialinecolor}{rgb}{0.000000, 0.000000, 0.000000}
\pgfsetstrokecolor{dialinecolor}
\pgfpathellipse{\pgfpoint{6.573437\du}{8.535937\du}}{\pgfpoint{1.464062\du}{0\du}}{\pgfpoint{0\du}{1.464062\du}}
\pgfusepath{stroke}
\pgfsetlinewidth{0.00000\du}
\pgfsetdash{}{0pt}
\pgfsetdash{}{0pt}
\pgfsetbuttcap
\pgfsetmiterjoin
\pgfsetlinewidth{0.00000\du}
\pgfsetbuttcap
\pgfsetmiterjoin
\pgfsetdash{}{0pt}
\definecolor{dialinecolor}{rgb}{1.000000, 1.000000, 1.000000}
\pgfsetfillcolor{dialinecolor}
\pgfpathellipse{\pgfpoint{12.519062\du}{8.609062\du}}{\pgfpoint{1.464062\du}{0\du}}{\pgfpoint{0\du}{1.464062\du}}
\pgfusepath{fill}
\definecolor{dialinecolor}{rgb}{0.000000, 0.000000, 0.000000}
\pgfsetstrokecolor{dialinecolor}
\pgfpathellipse{\pgfpoint{12.519062\du}{8.609062\du}}{\pgfpoint{1.464062\du}{0\du}}{\pgfpoint{0\du}{1.464062\du}}
\pgfusepath{stroke}
\pgfsetbuttcap
\pgfsetmiterjoin
\pgfsetdash{}{0pt}
\definecolor{dialinecolor}{rgb}{0.000000, 0.000000, 0.000000}
\pgfsetstrokecolor{dialinecolor}
\pgfpathellipse{\pgfpoint{12.519062\du}{8.609062\du}}{\pgfpoint{1.464062\du}{0\du}}{\pgfpoint{0\du}{1.464062\du}}
\pgfusepath{stroke}
\pgfsetlinewidth{0.00000\du}
\pgfsetdash{}{0pt}
\pgfsetdash{}{0pt}
\pgfsetbuttcap
{
\definecolor{dialinecolor}{rgb}{0.000000, 0.000000, 0.000000}
\pgfsetfillcolor{dialinecolor}
\definecolor{dialinecolor}{rgb}{0.000000, 0.000000, 0.000000}
\pgfsetstrokecolor{dialinecolor}
\draw (8.037500\du,8.65000\du)--(11.055000\du,8.65000\du);
}
\pgfsetlinewidth{0.00000\du}
\pgfsetdash{}{0pt}
\pgfsetdash{}{0pt}
\pgfsetbuttcap
{
\definecolor{dialinecolor}{rgb}{0.000000, 0.000000, 0.000000}
\pgfsetfillcolor{dialinecolor}
\definecolor{dialinecolor}{rgb}{0.000000, 0.000000, 0.000000}
\pgfsetstrokecolor{dialinecolor}
\draw (2.106197\du,8.65000\du)--(5.109375\du,8.65000\du);
}
\pgfsetlinewidth{0.00000\du}
\pgfsetdash{}{0pt}
\pgfsetdash{}{0pt}
\pgfsetbuttcap
{
\definecolor{dialinecolor}{rgb}{0.000000, 0.000000, 0.000000}
\pgfsetfillcolor{dialinecolor}
\definecolor{dialinecolor}{rgb}{0.000000, 0.000000, 0.000000}
\pgfsetstrokecolor{dialinecolor}
\draw (6.600000\du,7.071875\du)--(6.600000\du,3.950000\du);
}
\pgfsetlinewidth{0.00000\du}
\pgfsetdash{}{0pt}
\pgfsetdash{}{0pt}
\pgfsetbuttcap
{
\definecolor{dialinecolor}{rgb}{0.000000, 0.000000, 0.000000}
\pgfsetfillcolor{dialinecolor}
\definecolor{dialinecolor}{rgb}{0.000000, 0.000000, 0.000000}
\pgfsetstrokecolor{dialinecolor}
\draw (13.983125\du,8.65000\du)--(15.950000\du,8.650000\du);
}
\definecolor{dialinecolor}{rgb}{0.000000, 0.000000, 0.000000}
\pgfsetstrokecolor{dialinecolor}
\node[anchor=west] at (3.450000\du,8.000000\du){$n_1$};
\definecolor{dialinecolor}{rgb}{0.000000, 0.000000, 0.000000}
\pgfsetstrokecolor{dialinecolor}
\node[anchor=west] at (7.050000\du,5.400000\du){$n_2$};
\definecolor{dialinecolor}{rgb}{0.000000, 0.000000, 0.000000}
\pgfsetstrokecolor{dialinecolor}
\node[anchor=west] at (9.250000\du,8.000000\du){$n_3$};
\definecolor{dialinecolor}{rgb}{0.000000, 0.000000, 0.000000}
\pgfsetstrokecolor{dialinecolor}
\node[anchor=west] at (14.800000\du,8.000000\du){$p_3$};
\definecolor{dialinecolor}{rgb}{0.000000, 0.000000, 0.000000}
\pgfsetstrokecolor{dialinecolor}
\node[anchor=west] at (5.83437\du,8.65000\du){$\ten{A}$};
\definecolor{dialinecolor}{rgb}{0.000000, 0.000000, 0.000000}
\pgfsetstrokecolor{dialinecolor}
\node[anchor=west] at (11.91062\du,8.65000\du){$\mat{U}$};
\end{tikzpicture}
\caption{3-mode product between a 3-way tensor $\ten{A}$ and matrix $\mat{U}$.}
\label{fig:kmodeprod}
\end{center}
\end{figure}

Tensors are multi-dimensional arrays that are higher order generalization of vectors (first-order tensors) and matrices (second-order tensors). A $d$th-order or $d$-way tensor is denoted as $\ten{A}\in\mathbb{R}^{n_1\times n_2 \times \cdots \times n_d}$ and the element of $\ten{A}$ by $a_{i_1i_2...i_d}$, where 1$\le$ $i_k$ $\le$ $n_k$, $k = 1,2,\ldots,d$. The numbers $n_1, n_2,\ldots,n_d$ are called the dimensions of the tensor $\ten{A}$.
We use boldface capital calligraphic letters $\ten{A}$, $\ten{B}$, \ldots to denote tensors, boldface capital letters $\mat{A}$, $\mat{B}$, \ldots to denote matrices, boldface letters $\mat{a}$, $\mat{b}$, \ldots to denote vectors, and roman letters $a$, $b$, \ldots to denote scalars. $\mat{A}^T$ and $\mat{a}^T$ are the transpose of a matrix $\mat{A}$ and a vector $\mat{a}$. The unit matrix of order $n$ is denoted $\mat{I}_n$. An intuitive and useful graphical representation of scalars, vectors, matrices and tensors is depicted in Figure~\ref{fig:graphical}. The unconnected edges, also called free legs, are the indices of the array. Therefore scalars have no unconnected edge, while matrices have $2$ unconnected edges. We will mainly employ these graphical representations to visualize the tensor networks and operations in the following sections whenever possible and refer to~\cite{orus2014practical} for more details. We now briefly introduce some important tensor operations.
\begin{definition}(Tensor $k$-mode product): The $k$-mode product of a tensor $\ten{A}\in\mathbb{R}^{n_1\times\cdots \times n_k\times\cdots\times n_d}$ with a matrix $\mat{U}\in\mathbb{R}^{p_k\times n_k}$ is denoted as $\ten{B}= \ten{A}\times _k \mat{U}$ and defined by 
\begin{align}
\nonumber \ten{B}(i_1,\ldots,i_{k-1},j,i_{k+1},\ldots,i_d)&=\\
\nonumber  \sum\limits_{i_k=1}^{n_k}  \mat{U}(j,i_k)\, \ten{A}(i_1,\ldots,i_k,\ldots,i_d),
\end{align}
where $\ten{B}\in\mathbb{R}^{n_1\times\cdots \times n_{k-1}\times p_k \times n_{k+1}\times\cdots\times n_d}$.
\end{definition}
The graphical representation of a $3$-mode product between a third-order tensor $\ten{A}$ and a matrix $\mat{U}$ is shown in Figure~\ref{fig:kmodeprod}, where the summation over the $i_3$ index is indicated by the connected edge.
\begin{definition}(Reshaping) Reshaping is another often used tensor operation. Employing $MATLAB$ notation, ``$\textrm{reshape}(\ten{A},[m_1, m_2,\ldots,m_d] )$" reshapes the tensor $\ten{A}$ into another tensor with dimensions $m_1$, $m_2$, $\ldots, m_d$. The total number of elements of the tensor $\ten{A}$ must be $\prod_{k=1}^d m_k$.
\end{definition}
\begin{definition}(Vectorization) Vectorization is a special reshaping operation that reshapes a tensor $\ten{A}$ into a column vector, denoted as $\textrm{vec}(\ten{A})$.
\end{definition}
\begin{definition}(Tensor inner product)
For two tensors $\ten{A},\ten{B} \in \mathbb{R}^{n_1 \times n_2 \times \cdots \times n_d}$, their inner product $\langle \ten{A},\ten{B} \rangle$ is defined as
\begin{align*}
\langle \ten{A},\ten{B} \rangle &= \sum\limits_{i_1=1}^{n_1} \sum\limits_{i_2=1}^{n_2}\cdots \sum\limits_{i_d=1}^{n_d} a_{i_1,i_2,\cdots,i_d} b_{i_1,i_2,\cdots,i_d}.
\end{align*}
\end{definition}
\begin{definition}(Frobenius norm) The Frobenius norm of a tensor $\ten{A}\in\mathbb{R}^{n_1\times n_2 \times \cdots\times n_d}$ is defined as $||\ten{A}||_F=\sqrt{\langle \ten{A},\ten{A}\rangle}$.
\end{definition}

\subsection{Tensor Decompositions}
Here we introduce two related tensor decomposition methods, namely, the rank-1 tensor decomposition used in STM and the tensor train (TT) decomposition used in STTM.
\subsubsection{Tensor Rank-1 Decomposition}
A $d$-way tensor $\ten{A}\in\mathbb{R}^{n_1\times n_2\times\cdots\times n_d}$ is rank-1 if it can be written as the outer product of $d$ vectors
\begin{align}
\ten{A}=\mat{a}^{(1)} \circ \mat{a}^{(2)}\circ \cdots \circ\mat{a}^{(d)},
\end{align}
where  $\circ$ denotes the vector outer product, and each element in $\ten{A}$ is the product of the corresponding vector elements:
\begin{align}
\nonumber \ten{A}(i_1,\ldots,i_d)=\mat{a}^{(1)}(i_1)\mat{a}^{(2)}(i_2)\cdots  \mat{a}^{(d)}(i_d).
\end{align}
Storing the component vectors $\mat{a}^{(1)}, \ldots, \mat{a}^{(d)}$ instead of the whole tensor $\ten{A}$ significantly reduces the required number of storage elements. However, a rank-1 tensor is rare in real-world applications, so that a rank-1 approximation to a general tensor usually results in unacceptably large approximation errors. This calls for a more general and powerful tensor approximation, for which the TT decomposition serves as a particularly suitable choice.

\subsubsection{Tensor Train Decomposition}
\label{subsec:TT}
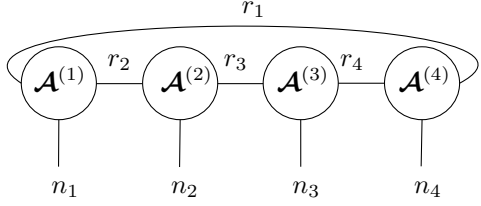
\begin{figure}[t]
\begin{center}
\ifx\du\undefined
  \newlength{\du}
\fi
\setlength{\du}{4\unitlength}
\begin{tikzpicture}
\pgftransformxscale{1.000000}
\pgftransformyscale{-1.0000}
\definecolor{dialinecolor}{rgb}{0.000000, 0.000000, 0.000000}
\pgfsetstrokecolor{dialinecolor}
\definecolor{dialinecolor}{rgb}{1.000000, 1.000000, 1.000000}
\pgfsetfillcolor{dialinecolor}
\definecolor{dialinecolor}{rgb}{1.000000, 1.000000, 1.000000}
\pgfsetfillcolor{dialinecolor}
\pgfpathellipse{\pgfpoint{10.273273\du}{9.997200\du}}{\pgfpoint{3.550000\du}{0\du}}{\pgfpoint{0\du}{3.450000\du}}
\pgfusepath{fill}
\pgfsetlinewidth{0.100000\du}
\pgfsetdash{}{0pt}
\pgfsetdash{}{0pt}
\definecolor{dialinecolor}{rgb}{0.000000, 0.000000, 0.000000}
\pgfsetstrokecolor{dialinecolor}
\pgfpathellipse{\pgfpoint{10.273273\du}{9.997200\du}}{\pgfpoint{3.550000\du}{0\du}}{\pgfpoint{0\du}{3.450000\du}}
\pgfusepath{stroke}
\pgfsetlinewidth{0.100000\du}
\pgfsetdash{}{0pt}
\pgfsetdash{}{0pt}
\pgfsetbuttcap
{
\definecolor{dialinecolor}{rgb}{0.000000, 0.000000, 0.000000}
\pgfsetfillcolor{dialinecolor}
\definecolor{dialinecolor}{rgb}{0.000000, 0.000000, 0.000000}
\pgfsetstrokecolor{dialinecolor}
\draw (13.823273\du,10.036600\du)--(18.196473\du,10.036600\du);
}
\definecolor{dialinecolor}{rgb}{0.000000, 0.000000, 0.000000}
\pgfsetstrokecolor{dialinecolor}
\node[anchor=west] at (8.7\du,20\du){$n_1$};
\definecolor{dialinecolor}{rgb}{1.000000, 1.000000, 1.000000}
\pgfsetfillcolor{dialinecolor}
\pgfpathellipse{\pgfpoint{21.746473\du}{9.997200\du}}{\pgfpoint{3.550000\du}{0\du}}{\pgfpoint{0\du}{3.450000\du}}
\pgfusepath{fill}
\pgfsetlinewidth{0.100000\du}
\pgfsetdash{}{0pt}
\pgfsetdash{}{0pt}
\definecolor{dialinecolor}{rgb}{0.000000, 0.000000, 0.000000}
\pgfsetstrokecolor{dialinecolor}
\pgfpathellipse{\pgfpoint{21.746473\du}{9.997200\du}}{\pgfpoint{3.550000\du}{0\du}}{\pgfpoint{0\du}{3.450000\du}}
\pgfusepath{stroke}
\pgfsetlinewidth{0.100000\du}
\pgfsetdash{}{0pt}
\pgfsetdash{}{0pt}
\pgfsetbuttcap
{
\definecolor{dialinecolor}{rgb}{0.000000, 0.000000, 0.000000}
\pgfsetfillcolor{dialinecolor}
\definecolor{dialinecolor}{rgb}{0.000000, 0.000000, 0.000000}
\pgfsetstrokecolor{dialinecolor}
\draw (21.746473\du,13.447200\du)--(21.691773\du,17.927200\du);
}
\definecolor{dialinecolor}{rgb}{0.000000, 0.000000, 0.000000}
\pgfsetstrokecolor{dialinecolor}
\node[anchor=west] at (20.\du,20\du){$n_2$};
\definecolor{dialinecolor}{rgb}{1.000000, 1.000000, 1.000000}
\pgfsetfillcolor{dialinecolor}
\pgfpathellipse{\pgfpoint{33.147073\du}{9.997200\du}}{\pgfpoint{3.550000\du}{0\du}}{\pgfpoint{0\du}{3.450000\du}}
\pgfusepath{fill}
\pgfsetlinewidth{0.100000\du}
\pgfsetdash{}{0pt}
\pgfsetdash{}{0pt}
\definecolor{dialinecolor}{rgb}{0.000000, 0.000000, 0.000000}
\pgfsetstrokecolor{dialinecolor}
\pgfpathellipse{\pgfpoint{33.147073\du}{9.997200\du}}{\pgfpoint{3.550000\du}{0\du}}{\pgfpoint{0\du}{3.450000\du}}
\pgfusepath{stroke}
\pgfsetlinewidth{0.100000\du}
\pgfsetdash{}{0pt}
\pgfsetdash{}{0pt}
\pgfsetbuttcap
{
\definecolor{dialinecolor}{rgb}{0.000000, 0.000000, 0.000000}
\pgfsetfillcolor{dialinecolor}
\definecolor{dialinecolor}{rgb}{0.000000, 0.000000, 0.000000}
\pgfsetstrokecolor{dialinecolor}
\draw (36.697073\du,9.997200\du)--(41.070273\du,9.997200\du);
}
\pgfsetlinewidth{0.100000\du}
\pgfsetdash{}{0pt}
\pgfsetdash{}{0pt}
\pgfsetbuttcap
{
\definecolor{dialinecolor}{rgb}{0.000000, 0.000000, 0.000000}
\pgfsetfillcolor{dialinecolor}
\definecolor{dialinecolor}{rgb}{0.000000, 0.000000, 0.000000}
\pgfsetstrokecolor{dialinecolor}
\draw (25.296473\du,10.036600\du)--(29.597073\du,10.036600\du);
}
\definecolor{dialinecolor}{rgb}{0.000000, 0.000000, 0.000000}
\pgfsetstrokecolor{dialinecolor}
\node[anchor=west] at (31.5\du,20\du){$n_3$};
\definecolor{dialinecolor}{rgb}{1.000000, 1.000000, 1.000000}
\pgfsetfillcolor{dialinecolor}
\pgfpathellipse{\pgfpoint{44.620273\du}{9.997200\du}}{\pgfpoint{3.550000\du}{0\du}}{\pgfpoint{0\du}{3.450000\du}}
\pgfusepath{fill}
\pgfsetlinewidth{0.100000\du}
\pgfsetdash{}{0pt}
\pgfsetdash{}{0pt}
\definecolor{dialinecolor}{rgb}{0.000000, 0.000000, 0.000000}
\pgfsetstrokecolor{dialinecolor}
\pgfpathellipse{\pgfpoint{44.620273\du}{9.997200\du}}{\pgfpoint{3.550000\du}{0\du}}{\pgfpoint{0\du}{3.450000\du}}
\pgfusepath{stroke}
\definecolor{dialinecolor}{rgb}{0.000000, 0.000000, 0.000000}
\pgfsetstrokecolor{dialinecolor}
\node[anchor=west] at (43\du,20\du){$n_4$};
\definecolor{dialinecolor}{rgb}{0.000000, 0.000000, 0.000000}
\pgfsetstrokecolor{dialinecolor}
\node[anchor=west] at (7.\du,9.554100\du){$\ten{A}^{(1)}$};
\definecolor{dialinecolor}{rgb}{0.000000, 0.000000, 0.000000}
\pgfsetstrokecolor{dialinecolor}
\node[anchor=west] at (19.\du,9.554100\du){$\ten{A}^{(2)}$};
\definecolor{dialinecolor}{rgb}{0.000000, 0.000000, 0.000000}
\pgfsetstrokecolor{dialinecolor}
\node[anchor=west] at (30.\du,9.554100\du){$\ten{A}^{(3)}$};
\definecolor{dialinecolor}{rgb}{0.000000, 0.000000, 0.000000}
\pgfsetstrokecolor{dialinecolor}
\node[anchor=west] at (41.5\du,9.554100\du){$\ten{A}^{(4)}$};
\definecolor{dialinecolor}{rgb}{0.000000, 0.000000, 0.000000}
\pgfsetstrokecolor{dialinecolor}
\node[anchor=west] at (26.697173\du,3.\du){$r_1$};
\definecolor{dialinecolor}{rgb}{0.000000, 0.000000, 0.000000}
\pgfsetstrokecolor{dialinecolor}
\node[anchor=west] at (14\du,8\du){$r_2$};
\definecolor{dialinecolor}{rgb}{0.000000, 0.000000, 0.000000}
\pgfsetstrokecolor{dialinecolor}
\node[anchor=west] at (25\du,8\du){$r_3$};
\definecolor{dialinecolor}{rgb}{0.000000, 0.000000, 0.000000}
\pgfsetstrokecolor{dialinecolor}
\node[anchor=west] at (36\du,8\du){$r_4$};
\pgfsetlinewidth{0.100000\du}
\pgfsetdash{}{0pt}
\pgfsetdash{}{0pt}
\pgfsetmiterjoin
\pgfsetbuttcap
{
\definecolor{dialinecolor}{rgb}{0.000000, 0.000000, 0.000000}
\pgfsetfillcolor{dialinecolor}
\definecolor{dialinecolor}{rgb}{0.000000, 0.000000, 0.000000}
\pgfsetstrokecolor{dialinecolor}
\pgfpathmoveto{\pgfpoint{6.723273\du}{9.997200\du}}
\pgfpathcurveto{\pgfpoint{-4.669327\du}{2.594600\du}}{\pgfpoint{61.340273\du}{2.983600\du}}{\pgfpoint{48.170273\du}{9.997200\du}}
\pgfusepath{stroke}
}
\pgfsetlinewidth{0.100000\du}
\pgfsetdash{}{0pt}
\pgfsetdash{}{0pt}
\pgfsetbuttcap
{
\definecolor{dialinecolor}{rgb}{0.000000, 0.000000, 0.000000}
\pgfsetfillcolor{dialinecolor}
\definecolor{dialinecolor}{rgb}{0.000000, 0.000000, 0.000000}
\pgfsetstrokecolor{dialinecolor}
\draw (10.273273\du,13.447200\du)--(10.245607\du,17.890607\du);
}
\pgfsetlinewidth{0.100000\du}
\pgfsetdash{}{0pt}
\pgfsetdash{}{0pt}
\pgfsetbuttcap
{
\definecolor{dialinecolor}{rgb}{0.000000, 0.000000, 0.000000}
\pgfsetfillcolor{dialinecolor}
\definecolor{dialinecolor}{rgb}{0.000000, 0.000000, 0.000000}
\pgfsetstrokecolor{dialinecolor}
\draw (33.147073\du,13.447200\du)--(33.150000\du,17.900000\du);
}
\pgfsetlinewidth{0.100000\du}
\pgfsetdash{}{0pt}
\pgfsetdash{}{0pt}
\pgfsetbuttcap
{
\definecolor{dialinecolor}{rgb}{0.000000, 0.000000, 0.000000}
\pgfsetfillcolor{dialinecolor}
\definecolor{dialinecolor}{rgb}{0.000000, 0.000000, 0.000000}
\pgfsetstrokecolor{dialinecolor}
\draw (44.620273\du,13.447200\du)--(44.535607\du,17.860607\du);
}
\end{tikzpicture}
\end{center}
\caption{Tensor train decomposition of a 4-way tensor $\ten{A}$ into 3-way tensors $\ten{A}^{(1)},\ldots,\ten{A}^{(4)}$.}
\label{fig:TT}
\end{figure}

A TT decomposition~\cite{oseledets2011tensor} represents a $d$-way tensor $\ten{A}$ as $d$ third-order tensors $\ten{A}^{(1)}$, $\ten{A}^{(2)}$, \ldots , $\ten{A}^{(d)}$ such that a particular entry of $\ten{A}$ is written as the following matrix product
\begin{align}
\ten{A}(i_1,\ldots,i_d)&=\ten{A}^{(1)}(:,i_1,:)\cdots \ten{A}^{(d)}(:,i_d,:).
\label{eq:TT}
\end{align}
Each tensor $\ten{A}^{(k)}$, $k=1,\ldots,d$, is called a TT-core and has dimensions \mbox{$r_k \times n_k \times r_{k+1}$}. Storage of a tensor as a TT therefore reduces from $\prod_{i=1}^d\,n_i$ down to $\sum_{i=1}^d\,r_in_ir_{i+1}$. In order for the left-hand-side of \eqref{eq:TT} to be a scalar we require that $r_1=r_{d+1}=1$. The remaining $r_k$ values are called the TT-ranks. Figure~\ref{fig:TT} illustrates the TT-decomposition of a 4-way tensor $\ten{A}$, where the edges connecting the different circles indicate the matrix-matrix products of \eqref{eq:TT}.

\begin{definition}(Left orthogonal and right orthogonal TT-cores)
A TT-core $\ten{A}^{(k)} (1\leq k \leq d)$ is left orthogonal when reshaped into an $r_{k}n_k \times r_{k+1}$ matrix $\mat{A}$ we have that 
\begin{align}
\nonumber \mat{A}^T\mat{A}=\mat{I}_{r_{k+1}}.
\end{align}
Similarly, a TT-core $\ten{A}^{(k)}$ is right orthogonal when reshaped into an $r_{k}\times n_kr_{k+1}$ matrix $\mat{A}$ we have that 
\begin{align}
\nonumber \mat{A}\mat{A}^T=\mat{I}_{r_{k}}.
\end{align}  
\end{definition}
\begin{definition}(Site-$k$-mixed-canonical tensor train)
A tensor train is in site-$k$-mixed-canonical form~\cite{SCHOLLWOCK201196} when all TT-cores $\{\ten{A}^{(l)}\,|\,l=1,\ldots,k-1\}$ are left orthogonal and $\{\ten{A}^{(l)}\,|\, l=k+1,\ldots,d\}$ are right orthogonal.
\end{definition}
Turning a TT into its site-$k$-mixed-canonical form requires $d-1$ QR decompositions of the reshaped TT-cores. Changing $k$ in a site-$k$-mixed-canonical form to either $k-1$ or $k+1$ requires one QR factorization of $\ten{A}^{(k)}$. It can be shown that the Frobenius norm of a tensor $\ten{A}$ in a site-$k$-mixed-canonical form is easily computed from
\begin{align}
\nonumber {||\ten{A}||}_F^2&={||\ten{A}^{(k)}||}_F^2 = \textrm{vec}(\ten{A}^{(k)})^T \textrm{vec}(\ten{A}^{(k)}).
\end{align}

\subsection{Support Vector Machines}
We briefly introduce linear SVMs before discussing STMs. Assume we have a dataset~\mbox{$D$=\{$\mat{x_i}$, $y_i$\}$_{i=1}^M$} of $M$ labeled samples, where $\mat{x}_i \in \mathbb{R}^n$ are the samples or feature vectors with labels $y_i\in \{-1,1\}$. Learning a linear SVM is finding a discriminant hyperplane 
\begin{align}
  f(\mat{x})=\mat{w}^T\mat{x}+b
\label{eq:hyperplane}
\end{align}
that maximizes the margin between the two classes where $\mat{w}$ and $b$ are the weight vector and bias, respectively. In practice, the data are seldom linearly separable due to measurement noise. A more robust classifier can then be found by introducing the slack variables $\xi_1,\ldots,\xi_M$ and writing the learning problem as an optimization problem
\begin{align}
\label{eq:svmQP}
\nonumber \min_{\mat{w},b,\xi} 		&\quad \, \frac{1}{2} || \mat{w}||_F^2 +C\sum\limits_{i=1}^{M} \xi_i \\
\nonumber \textrm{subject to}  		&\quad \, y_i( \mat{w}^T \mat{x}_i+b )\ge 1-\xi_i,\\
									&\quad \, \xi_i\ge 0,\; i=1,\ldots ,M.
\end{align}
The parameter $C$ controls the trade-off between the size of the weight vector $\mat{w}$ and the size of the slack variables. It is common to solve the dual problem of~\eqref{eq:svmQP} with quadratic programming, especially when the feature size $n$ is larger than the sample size $M$.

\subsection{Support Tensor Machines}
\label{subsec:STM}
\begin{figure}[t]
\begin{center}
\ifx\du\undefined
  \newlength{\du}
\fi
\setlength{\du}{15\unitlength}
\begin{tikzpicture}
\pgftransformxscale{0.950000}
\pgftransformyscale{-0.950000}
\definecolor{dialinecolor}{rgb}{0.000000, 0.000000, 0.000000}
\pgfsetstrokecolor{dialinecolor}
\definecolor{dialinecolor}{rgb}{1.000000, 1.000000, 1.000000}
\pgfsetfillcolor{dialinecolor}
\pgfsetlinewidth{0.00000\du}
\pgfsetdash{}{0pt}
\pgfsetdash{}{0pt}
\pgfsetbuttcap
\pgfsetmiterjoin
\pgfsetlinewidth{0.00000\du}
\pgfsetbuttcap
\pgfsetmiterjoin
\pgfsetdash{}{0pt}
\definecolor{dialinecolor}{rgb}{1.000000, 1.000000, 1.000000}
\pgfsetfillcolor{dialinecolor}
\pgfpathellipse{\pgfpoint{5.050000\du}{10.450000\du}}{\pgfpoint{1.050000\du}{0\du}}{\pgfpoint{0\du}{1.050000\du}}
\pgfusepath{fill}
\definecolor{dialinecolor}{rgb}{0.000000, 0.000000, 0.000000}
\pgfsetstrokecolor{dialinecolor}
\pgfpathellipse{\pgfpoint{6.050000\du}{10.450000\du}}{\pgfpoint{1.050000\du}{0\du}}{\pgfpoint{0\du}{1.050000\du}}
\pgfusepath{stroke}
\pgfsetbuttcap
\pgfsetmiterjoin
\pgfsetdash{}{0pt}
\definecolor{dialinecolor}{rgb}{0.000000, 0.000000, 0.000000}
\pgfsetstrokecolor{dialinecolor}
\node[anchor=west] at (5.050000\du,10.450000\du){$f(\ten{X})$};
\definecolor{dialinecolor}{rgb}{0.000000, 0.000000, 0.000000}
\pgfsetstrokecolor{dialinecolor}
\node[anchor=west] at (7.30000\du,10.400000\du){=};
\pgfsetlinewidth{0.00000\du}
\pgfsetdash{}{0pt}
\pgfsetdash{}{0pt}
\pgfsetbuttcap
\pgfsetmiterjoin
\pgfsetlinewidth{0.00000\du}
\pgfsetbuttcap
\pgfsetmiterjoin
\pgfsetdash{}{0pt}
\definecolor{dialinecolor}{rgb}{1.000000, 1.000000, 1.000000}
\pgfsetfillcolor{dialinecolor}
\pgfpathellipse{\pgfpoint{11.842500\du}{10.382500\du}}{\pgfpoint{1.237500\du}{0\du}}{\pgfpoint{0\du}{1.237500\du}}
\pgfusepath{fill}
\definecolor{dialinecolor}{rgb}{0.000000, 0.000000, 0.000000}
\pgfsetstrokecolor{dialinecolor}
\pgfpathellipse{\pgfpoint{11.842500\du}{10.382500\du}}{\pgfpoint{1.237500\du}{0\du}}{\pgfpoint{0\du}{1.237500\du}}
\pgfusepath{stroke}
\pgfsetbuttcap
\pgfsetmiterjoin
\pgfsetdash{}{0pt}
\definecolor{dialinecolor}{rgb}{0.000000, 0.000000, 0.000000}
\pgfsetstrokecolor{dialinecolor}
\pgfpathellipse{\pgfpoint{11.842500\du}{10.382500\du}}{\pgfpoint{1.237500\du}{0\du}}{\pgfpoint{0\du}{1.237500\du}}
\pgfusepath{stroke}
\pgfsetlinewidth{0.00000\du}
\pgfsetdash{}{0pt}
\pgfsetdash{}{0pt}
\pgfsetbuttcap
{
\definecolor{dialinecolor}{rgb}{0.000000, 0.000000, 0.000000}
\pgfsetfillcolor{dialinecolor}
\definecolor{dialinecolor}{rgb}{0.000000, 0.000000, 0.000000}
\pgfsetstrokecolor{dialinecolor}
\draw (9.850000\du,8.650000\du)--(10.850000\du,9.650000\du);
}
\pgfsetlinewidth{0.00000\du}
\pgfsetdash{}{0pt}
\pgfsetdash{}{0pt}
\pgfsetbuttcap
{
\definecolor{dialinecolor}{rgb}{0.000000, 0.000000, 0.000000}
\pgfsetfillcolor{dialinecolor}
\definecolor{dialinecolor}{rgb}{0.000000, 0.000000, 0.000000}
\pgfsetstrokecolor{dialinecolor}
\draw (11.523634\du,7.906810\du)--(11.678124\du,9.106276\du);
}
\pgfsetlinewidth{0.00000\du}
\pgfsetdash{}{0pt}
\pgfsetdash{}{0pt}
\pgfsetbuttcap
{
\definecolor{dialinecolor}{rgb}{0.000000, 0.000000, 0.000000}
\pgfsetfillcolor{dialinecolor}
\definecolor{dialinecolor}{rgb}{0.000000, 0.000000, 0.000000}
\pgfsetstrokecolor{dialinecolor}
\draw (14.050000\du,8.650000\du)--(12.980711\du,9.860711\du);
}
\definecolor{dialinecolor}{rgb}{0.000000, 0.000000, 0.000000}
\pgfsetstrokecolor{dialinecolor}
\node[anchor=west] at (12.500000\du,8.200000\du){...};
\pgfsetlinewidth{0.00000\du}
\pgfsetdash{}{0pt}
\pgfsetdash{}{0pt}
\pgfsetbuttcap
\pgfsetmiterjoin
\pgfsetlinewidth{0.00000\du}
\pgfsetbuttcap
\pgfsetmiterjoin
\pgfsetdash{}{0pt}
\definecolor{dialinecolor}{rgb}{1.000000, 1.000000, 1.000000}
\pgfsetfillcolor{dialinecolor}
\pgfpathellipse{\pgfpoint{14.697500\du}{8.027500\du}}{\pgfpoint{0.937500\du}{0\du}}{\pgfpoint{0\du}{0.937500\du}}
\pgfusepath{fill}
\definecolor{dialinecolor}{rgb}{0.000000, 0.000000, 0.000000}
\pgfsetstrokecolor{dialinecolor}
\pgfpathellipse{\pgfpoint{14.697500\du}{8.027500\du}}{\pgfpoint{0.937500\du}{0\du}}{\pgfpoint{0\du}{0.937500\du}}
\pgfusepath{stroke}
\pgfsetbuttcap
\pgfsetmiterjoin
\pgfsetdash{}{0pt}
\definecolor{dialinecolor}{rgb}{0.000000, 0.000000, 0.000000}
\pgfsetstrokecolor{dialinecolor}
\pgfpathellipse{\pgfpoint{14.697500\du}{8.027500\du}}{\pgfpoint{0.937500\du}{0\du}}{\pgfpoint{0\du}{0.937500\du}}
\pgfusepath{stroke}
\definecolor{dialinecolor}{rgb}{0.000000, 0.000000, 0.000000}
\pgfsetstrokecolor{dialinecolor}
\node[anchor=west] at (11.242500\du,10.382500\du){$\ten{X}$};
\definecolor{dialinecolor}{rgb}{0.000000, 0.000000, 0.000000}
\pgfsetstrokecolor{dialinecolor}
\node[anchor=west] at (9.837500\du,8.212500\du){};
\definecolor{dialinecolor}{rgb}{0.000000, 0.000000, 0.000000}
\pgfsetstrokecolor{dialinecolor}
\node[anchor=west] at (9.242500\du,8.132500\du){};
\definecolor{dialinecolor}{rgb}{0.000000, 0.000000, 0.000000}
\pgfsetstrokecolor{dialinecolor}
\node[anchor=west] at (13.907500\du,8.027500\du){$\mat{w}^{(d)}$};
\definecolor{dialinecolor}{rgb}{0.000000, 0.000000, 0.000000}
\pgfsetstrokecolor{dialinecolor}
\node[anchor=west] at (15.950000\du,10.400000\du){+};
\pgfsetlinewidth{0.00000\du}
\pgfsetdash{}{0pt}
\pgfsetdash{}{0pt}
\pgfsetbuttcap
\pgfsetmiterjoin
\pgfsetlinewidth{0.00000\du}
\pgfsetbuttcap
\pgfsetmiterjoin
\pgfsetdash{}{0pt}
\definecolor{dialinecolor}{rgb}{1.000000, 1.000000, 1.000000}
\pgfsetfillcolor{dialinecolor}
\pgfpathellipse{\pgfpoint{18.505000\du}{10.445000\du}}{\pgfpoint{1.050000\du}{0\du}}{\pgfpoint{0\du}{1.050000\du}}
\pgfusepath{fill}
\definecolor{dialinecolor}{rgb}{0.000000, 0.000000, 0.000000}
\pgfsetstrokecolor{dialinecolor}
\pgfpathellipse{\pgfpoint{18.505000\du}{10.445000\du}}{\pgfpoint{1.050000\du}{0\du}}{\pgfpoint{0\du}{1.050000\du}}
\pgfusepath{stroke}
\pgfsetbuttcap
\pgfsetmiterjoin
\pgfsetdash{}{0pt}
\definecolor{dialinecolor}{rgb}{0.000000, 0.000000, 0.000000}
\pgfsetstrokecolor{dialinecolor}
\pgfpathellipse{\pgfpoint{18.505000\du}{10.445000\du}}{\pgfpoint{1.050000\du}{0\du}}{\pgfpoint{0\du}{1.050000\du}}
\pgfusepath{stroke}
\definecolor{dialinecolor}{rgb}{0.000000, 0.000000, 0.000000}
\pgfsetstrokecolor{dialinecolor}
\node[anchor=west] at (18.105000\du,10.445000\du){$b$};
\pgfsetlinewidth{0.00000\du}
\pgfsetdash{}{0pt}
\pgfsetdash{}{0pt}
\pgfsetbuttcap
\pgfsetmiterjoin
\pgfsetlinewidth{0.00000\du}
\pgfsetbuttcap
\pgfsetmiterjoin
\pgfsetdash{}{0pt}
\definecolor{dialinecolor}{rgb}{1.000000, 1.000000, 1.000000}
\pgfsetfillcolor{dialinecolor}
\pgfpathellipse{\pgfpoint{9.242500\du}{8.132500\du}}{\pgfpoint{0.937500\du}{0\du}}{\pgfpoint{0\du}{0.937500\du}}
\pgfusepath{fill}
\definecolor{dialinecolor}{rgb}{0.000000, 0.000000, 0.000000}
\pgfsetstrokecolor{dialinecolor}
\pgfpathellipse{\pgfpoint{9.242500\du}{8.132500\du}}{\pgfpoint{0.937500\du}{0\du}}{\pgfpoint{0\du}{0.937500\du}}
\pgfusepath{stroke}
\pgfsetbuttcap
\pgfsetmiterjoin
\pgfsetdash{}{0pt}
\definecolor{dialinecolor}{rgb}{0.000000, 0.000000, 0.000000}
\pgfsetstrokecolor{dialinecolor}
\pgfpathellipse{\pgfpoint{9.242500\du}{8.132500\du}}{\pgfpoint{0.937500\du}{0\du}}{\pgfpoint{0\du}{0.937500\du}}
\pgfusepath{stroke}
\pgfsetlinewidth{0.00000\du}
\pgfsetdash{}{0pt}
\pgfsetdash{}{0pt}
\pgfsetbuttcap
\pgfsetmiterjoin
\pgfsetlinewidth{0.00000\du}
\pgfsetbuttcap
\pgfsetmiterjoin
\pgfsetdash{}{0pt}
\definecolor{dialinecolor}{rgb}{1.000000, 1.000000, 1.000000}
\pgfsetfillcolor{dialinecolor}
\pgfpathellipse{\pgfpoint{11.397500\du}{6.927500\du}}{\pgfpoint{0.937500\du}{0\du}}{\pgfpoint{0\du}{0.937500\du}}
\pgfusepath{fill}
\definecolor{dialinecolor}{rgb}{0.000000, 0.000000, 0.000000}
\pgfsetstrokecolor{dialinecolor}
\pgfpathellipse{\pgfpoint{11.397500\du}{6.927500\du}}{\pgfpoint{0.937500\du}{0\du}}{\pgfpoint{0\du}{0.937500\du}}
\pgfusepath{stroke}
\pgfsetbuttcap
\pgfsetmiterjoin
\pgfsetdash{}{0pt}
\definecolor{dialinecolor}{rgb}{0.000000, 0.000000, 0.000000}
\pgfsetstrokecolor{dialinecolor}
\pgfpathellipse{\pgfpoint{11.397500\du}{6.927500\du}}{\pgfpoint{0.937500\du}{0\du}}{\pgfpoint{0\du}{0.937500\du}}
\pgfusepath{stroke}
\definecolor{dialinecolor}{rgb}{0.000000, 0.000000, 0.000000}
\pgfsetstrokecolor{dialinecolor}
\node[anchor=west] at (8.442500\du,8.132500\du){$\mat{w}^{(1)}$};
\definecolor{dialinecolor}{rgb}{0.000000, 0.000000, 0.000000}
\pgfsetstrokecolor{dialinecolor}
\node[anchor=west] at (10.597500\du,6.927500\du){$\mat{w}^{(2)}$};
\end{tikzpicture}
\end{center}
\caption{Graphical representation of an STM hyperplane function.}
\label{fig:STM}
\end{figure}
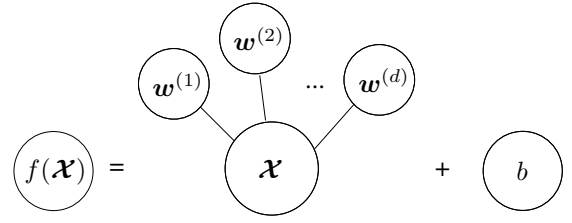

Suppose the input samples in the dataset \mbox{$D$=\{$\ten{X}_i$, $y_i$\}$_{i=1}^M$} are tensors $\ten{X}_i \in\mathbb{R}^{n_1\times n_2\times\cdots\times n_d}$. A linear STM extends a linear SVM by defining $d$ weight vectors $\mat{w}^{(i)} \in \mathbb{R}^{n_i}$ ($i=1,\ldots,d$) and rewriting~\eqref{eq:hyperplane} as
\begin{align}
f(\ten{X})&=\ten{X}\times_1 \mat{w}^{(1)} \times_2 \cdots \times_d \mat{w}^{(d)} + b.\label{eq:hyperplaneSTM}
\end{align}
The graphical representation of~\eqref{eq:hyperplaneSTM} is shown in Figure~\ref{fig:STM}. 
The tensor $\ten{X}$ is contracted along each of its modes with the weight vectors $\mat{w}^{(1)},\ldots,\mat{w}^{(d)}$, resulting in a scalar that is added to the bias $b$. The weight vectors of the STM are computed by the alternating projection optimization procedure, which comprises $d$ optimization problems. The main idea is to optimize each $\mat{w}^{(k)}$ in turn by fixing all weight vectors but $\mat{w}^{(k)}$. The $k$th optimization problem is
\begin{align}
\nonumber \min_{\mat{w}^{(k)},b,\xi} \quad &\frac{1}{2}\,\beta \, ||\mat{w}^{(k)}||_F^2 +C\sum\limits_{i=1}^{M} \xi_i\\
\nonumber \textrm{subject to} \quad &y_i(( \mat{w}^{(k)})^T\hat{\mat{x}}_i +b)\ge 1-\xi_i,\\
&\xi_i\ge 0,\; i=1,\ldots ,M,
\label{eq:stmQP}
\end{align}
where
\begin{align*}
 \beta =\prod_{1 \leq l \leq d}^{l \neq k} ||\mat{w}^{(l)}||_F^2\mbox{~~and~~}\hat{\mat{x}}_i =\ten{X}_i\prod_{1 \leq l \leq d}^{l\neq k}  \times_l \; \mat{w}^{(l)}.
\end{align*}
The optimization problem~\eqref{eq:stmQP} is equivalent to~\eqref{eq:svmQP} for the linear SVM problem. This implies that any SVM learning algorithm can also be used for the linear STM. Each of the weight vectors of the linear STM is updated consecutively until the loss function of~\eqref{eq:stmQP} converges. The convergence proof can be found in \cite[p.~14]{Tao2007}. Each single optimization problem in learning an STM requires the estimation of only a few weight parameters, which alleviates the overfitting problem when $M$ is relatively small. The weight tensor obtained from the outer product of the weight vectors
\begin{align}
\ten{W}&= \mat{w}^{(1)} \circ \mat{w}^{(2)} \circ \cdots \circ \mat{w}^{(d)}
\end{align}
is per definition rank-1 and allows us to rewrite~\eqref{eq:hyperplaneSTM} as
\begin{align}
f(\ten{X})&= \langle \ten{W},\ten{X} \rangle + b.
\end{align}
The constraint that $\ten{W}$ is a rank-1 tensor has a significant impact on the expressive power of the STM, resulting in an usually unsatisfactory classification accuracy for many real-world data. In this paper, we address this problem by representing $\ten{W}$ as a TT with prescribed TT-ranks.

\section{Support Tensor Train Machines}
\label{sec:STTM}
\begin{figure}[t]
\begin{center}
\input{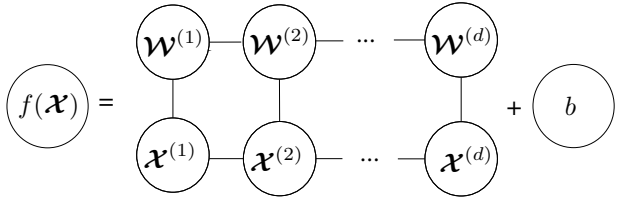}
\end{center}
\caption{Tensor graphical representation of an STTM hyperplane function.}
\label{fig:STTM}
\end{figure}

\begin{figure*}[h]
\begin{center}
\input{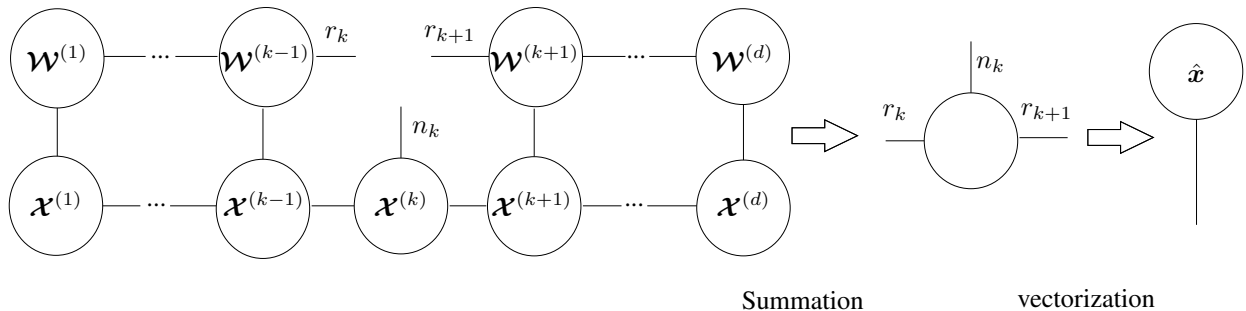}
\end{center}
\caption{The computation diagram of $\hat{\mat{x}}$.}
\label{fig:STTMxhat}
\end{figure*}

We first introduce our proposed STTM for binary classification, and then extend it to the multi-classification case. The graphical representation for tensors shown in Figure~\ref{fig:graphical} will be used to illustrate the different operations. As mentioned in Section~\ref{subsec:STM}, an STM suffers from its weak expressive power due to its rank-1 weight tensor $\ten{W}$. To this end, the proposed STTM replaces the rank-1 weight tensor by a TT with prescribed TT-ranks. Moreover, most real-world data contains redundancies and uninformative parts. Based on this knowledge, STTM also utilizes a TT decomposition to approximate the original data tensor as to alleviate the overfitting problem even further. The conversion of the training sample to a TT can be done using the TT-SVD algorithm~\cite[p.~2301]{oseledets2011tensor}, which allows the user to determine the relative error of the approximation. A graphical representation of the STTM hyperplane equation is shown in Figure~\ref{fig:STTM}. Both the data tensor $\ten{X}$ and the weight tensor $\ten{W}$ are represented by TTs and the summations correspond to computing the inner product $\langle \ten{X}, \ten{W} \rangle$.
The TT-cores $\ten{W}^{(1)}$, $\ten{W}^{(2)}$, $\ldots$, $\ten{W}^{(d)}$ are also computed using an alternating projection optimization procedure~\cite{tao2005supervised}, namely iteratively fixing $d-1$ TT-cores and updating the remaining core until convergence. This updating occurs in a ``looping'' fashion, whereby we first update $\ten{W}^{(1)}$ and proceed towards $\ten{W}^{(d)}$. After updating $\ten{W}^{(d)}$, we go around the loop and update $\ten{W}^{(1)}$.
Suppose we want to update $\ten{W}^{(k)}$. First, the TT of the weight tensor $\ten{W}$ is brought into site-$k$-mixed-canonical form. From Section~\ref{subsec:TT}, the norm of the whole weight tensor is located in the $\ten{W}^{(k)}$ TT-core. In order to reformulate the optimization problem~\eqref{eq:stmQP} in terms of the unknown core $\ten{W}^{(k)}$, we first need to re-express the inner product $\langle \ten{X},\ten{W} \rangle$ in terms of $\ten{W}^{(k)}$ as $\textrm{vec}(\ten{W}^{(k)})^T\hat{\mat{x}}$. The vector $\hat{\mat{x}}$ is obtained by summing over the tensor network for $\langle \ten{W},\ten{X} \rangle$ depicted in Figure~\ref{fig:STTM} with the TT-core $\ten{W}^{(k)}$ removed and vectorizing the resulting 3-way tensor. These two computational steps to compute $\hat{\mat{x}}$ are graphically depicted in Figure~\ref{fig:STTMxhat}. The STTM hyperplane function can then be rewritten as $\textrm{vec}(\ten{W}^{(k)})^T\hat{\mat{x}}+b$, so that $\ten{W}^{(k)}$ can be updated from the following optimization problem
\begin{align}
\nonumber \min\limits_{\ten{W}^{(k)},b,\xi} \quad &\frac{1}{2}||\ten{W}^{(k)}||_F^2+C\sum\limits_{i=1}^{M} \xi_i\\
\nonumber \textrm{subject to}\quad  &y_i(\textrm{vec}(\ten{W}^{(k)})^T\hat{\mat{x}}_i+b)\ge1-\xi_i,\\
&\xi_i\ge 0,\; i=1,\ldots ,M,
\label{eq:sttmQP}
\end{align}
using any computational method for standard SVMs. Suppose now that the next TT-core to be updated is $\ten{W}^{(k+1)}$. The new TT for $\ten{W}$ then needs to be put into site-$(k+1)$-mixed-canonical form, which can be achieved by reshaping the new $\ten{W}^{(k)}$ into an $r_kn_k\times r_{k+1}$ matrix $\mat{W}^{(k)}$ and computing its thin QR decomposition
\begin{align*}
\mat{W}^{(k)} &= \mat{Q} \;\mat{R},
\end{align*}
where $\mat{Q}$ is a $r_kn_k \times r_{k+1}$ matrix with orthogonal columns and $\mat{R}$ is an $r_{k+1}\times r_{k+1}$ upper triangular matrix. Updating the tensors $\ten{W}^{(k)},\ten{W}^{(k+1)} $ as
\begin{align*}
\ten{W}^{(k)} &:= \textrm{reshape}(\mat{Q},[r_k,n_k,r_{k+1}]),\\
\ten{W}^{(k+1)} &:= \ten{W}^{(k+1)} \times_1 \mat{R},
\end{align*}
results in a site-$(k+1)$-mixed-canonical form for $\ten{W}$. An optimization problem similar to~\eqref{eq:sttmQP} can then be derived for $\ten{W}^{(k+1)}$.
\begin{figure}[h]
\centering
\includegraphics[width=3.5in]{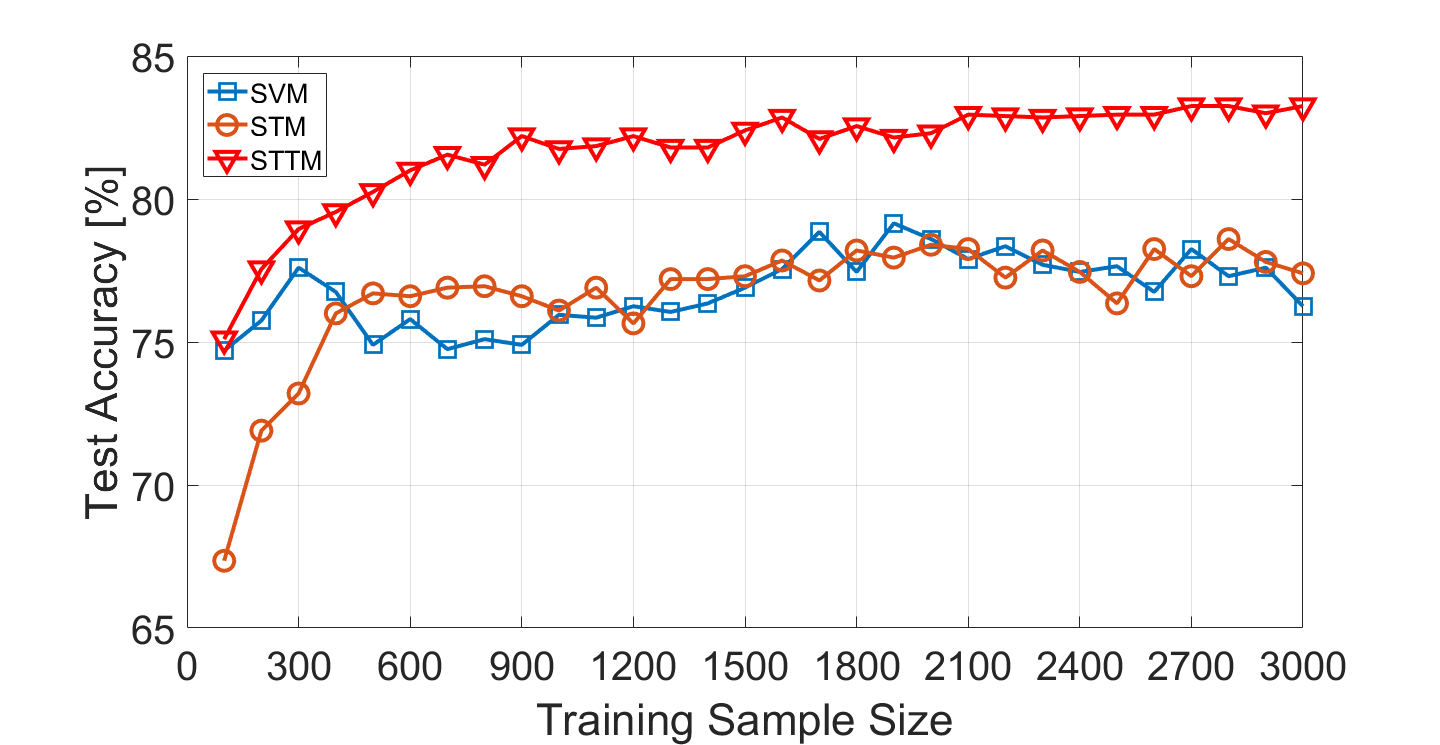}
\caption{Test accuracy of SVM, STM and STTM trained with different sample sizes.}
\label{fig:experiment1_1}
\end{figure}
The training algorithm of the STTM is summarized as pseudo-codes in Algorithm \ref{alg:STTM}. The TT-cores for the weight tensor $\ten{W}$ are initialized randomly. Bringing this TT into site-1-mixed-canonical form can then be done by applying the QR decomposition step starting from $\ten{W}^{(2)}$ and proceeding towards $\ten{W}^{(d)}$. The final $\mat{R}$ factor is absorbed by $\ten{W}^{(1)}$, which brings the TT into site-1-mixed-canonical form. The termination criterion in line 4 can be a maximum number of loops and/or when the training error falls below a user-defined threshold.  
\begin{algorithm}[t]
	\renewcommand{\algorithmicrequire}{\textbf{Input:}}
	\renewcommand{\algorithmicensure}{\textbf{Output:}}
	\caption{STTM Algorithm}
	\label{alg:STTM}
	\begin{algorithmic}[1]
		\REQUIRE TT-ranks $r_2, \ldots, r_{d}$ of $\ten{W}^{(1)}, \ten{W}^{(2)}, \ldots, \ten{W}^{(d)}$; Training dataset $\{\ten{X}_i\in\mathbb{R}^{n_1\times \cdots\times n_d}$, $y_i\in \{-1, 1 \}$\}$_{i=1}^M$;  Relative error $\epsilon$ of TT approximation of $\ten{X}$.   
		\ENSURE The TT-cores $\ten{W}^{(1)}, \ten{W}^{(2)}, \ldots, \ten{W}^{(d)}$; The bias $b$.
        \vspace{1ex}
        \STATE Initialize $\ten{W}^{(k)} \in \mathbb{R}^{r_k\times n_k\times r_{k+1}}$ as a random/prescribed 3-way tensor for $k=1, 2, \ldots, d$. 		
		\STATE Compute the TT approximation of training samples $\{\ten{X}_i\}_{i=1}^M$ with relative error $\epsilon$ using TT-SVD.
        \STATE Cast $\ten{W}$ into the site-1-mixed-canonical TT form.        
        \WHILE{termination criterion not satisfied}
        \FOR{$k=1,\ldots,d$}
        \STATE $\ten{W}^{(k)}, b \gets $ Solve optimization problem~\eqref{eq:sttmQP}.
        \STATE $\mat{W}^{(k)} \gets \textrm{reshape}(\ten{W}^{(k)},[r_kn_k,r_{k+1}])$.
        \STATE Compute thin QR decomposition $\mat{W}^{(k)}=\mat{Q}\mat{R}$.
        \STATE $\ten{W}^{(k)} \gets \textrm{reshape}(\mat{Q},[r_k,n_k,r_{k+1}])$.
        \STATE {$\ten{W}^{(k+1)} \gets \ten{W}^{(k+1)} \times_1 \mat{R}$.  \hfill \%$\ten{W}^{(d+1)}=\ten{W}^{(1)}$}
        \ENDFOR
        \ENDWHILE   		
	\end{algorithmic}%
\end{algorithm}%
To extend the binary classification STTM to an $L$-class classification STTM, we employ the one-versus-one strategy due to accuracy considerations~\cite{hsu2002comparison}. Specifically, we construct $L(L-1)/2$ binary classification STTMs, where each STTM is trained on data samples from two classes. The label of a test sample is then predicted by a majority voting strategy.


\section{Experiments}
\label{sec:experiments}
We present two experiments to show the superiority of the proposed STTM over standard SVM and STM in terms of classification accuracy. All experiments are implemented in MATLAB on an Intel i5 3.2GHz desktop with 16GB RAM. Note that STM and STTM can separate their overall optimization problem into $d$ standard SVM problems, namely,~\eqref{eq:stmQP} and~\eqref{eq:sttmQP}, respectively. For fair comparison, we employ the MATLAB built-in SVM solver {\tt fitcsvm} to get the solution for standard SVM, STM and STTM. When calling {\tt fitcsvm}, we select a linear kernel with default parameters and set the outlier fraction as 2\% for all experiments. The runtimes for training SVM, STM and STTM are very similar, all ranging from seconds to minutes depending on the training sample batch size. For this reason the classification accuracy is used to compare the different training algorithms with one another in the following experiments.

\subsection{CIFAR-10 Binary Classification}
Here we demonstrate three different aspects of the proposed STTM method: a comparison of its test accuracy versus SVMs and STMs, the influence of the TT-rank on the test accuracy, and the necessity of using the site-$k$-mixed-canonical form in Algorithm~\ref{alg:STTM}. 
\subsubsection{Classification}
The CIFAR-10 database~\cite{krizhevsky2009learning} is used in this binary classification experiment, which consists of 60k $32\times 32$ color images from $10$ classes, with $6000$ images each. The {\tt airplane} and {\tt automobile} classes were arbitrarily chosen to compare the test accuracy of the proposed STTM with SVM and STM. The first $3000$ samples of both classes were used for training while the rest were used as test data. Vectorizing the data samples results in a feature dimension of $3072$, which may lead to overfitting when the training sample size is much smaller. To verify the effectiveness of STTM with different training sample sizes, we divided the $3000$ training samples into $30$ experiments of varying sample batch sizes, namely $100$, $200$,$\ldots$, $2900$, $3000$. For each batch size we trained a standard SVM, STM and STTM. Prior to training the STTM, each data sample was converted into a TT of $3$ TT-cores with dimensions $n_1=n_2=32,n_3=3$ and $\epsilon=10^{-2}$. The TT-ranks of the weight TT were fixed to $r_1=r_4=1,r_3=3$ and different experiment runs were performed where $r_2$ varied from $2$ to $32$. The best test accuracy, defined as the percentage of correctly classified test samples, among these $31$ TT-ranks are compared with the test accuracy of both the SVM and STM methods in Figure~\ref{fig:experiment1_1}. STTM achieves the best test accuracy in all the sample batch sizes, while STM sometimes performs worse than a standard SVM, especially when the batch size is below $400$. The limitation on the performance of the STM is probably due to the poor expressive power of the rank-1 weight tensor. A batch size of $500$ samples suffices for the STTM to achieve the best test accuracy of the standard SVM over all sample sizes, which demonstrates the superiority of STTM at fewer training samples.%

\begin{figure}[t]
\centering
\includegraphics[width=3.5in]{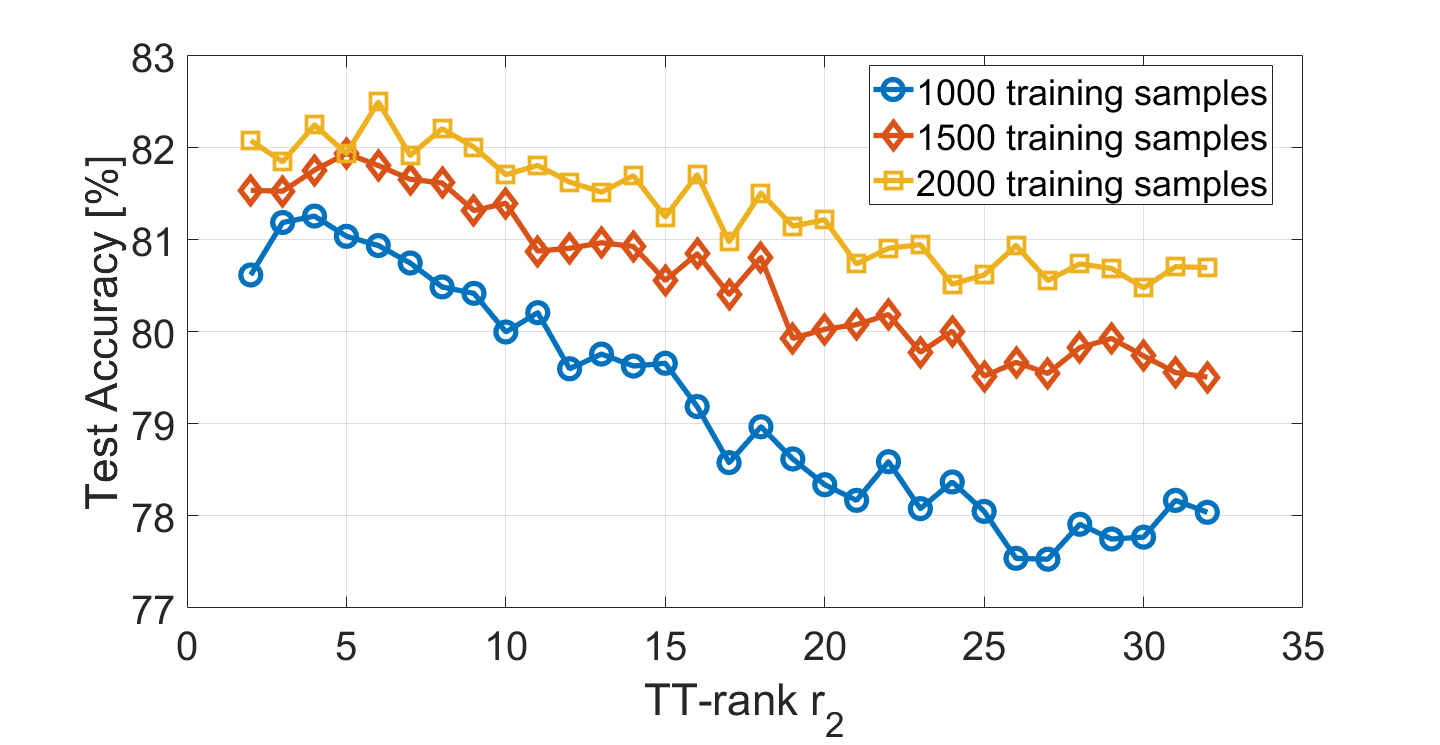}
\caption{Test accuracy of STTM on different TT-rank $r_2$.}
\label{fig:experiment1_2}
\end{figure}
\begin{figure}[t]
\centering
\includegraphics[width=3.5in]{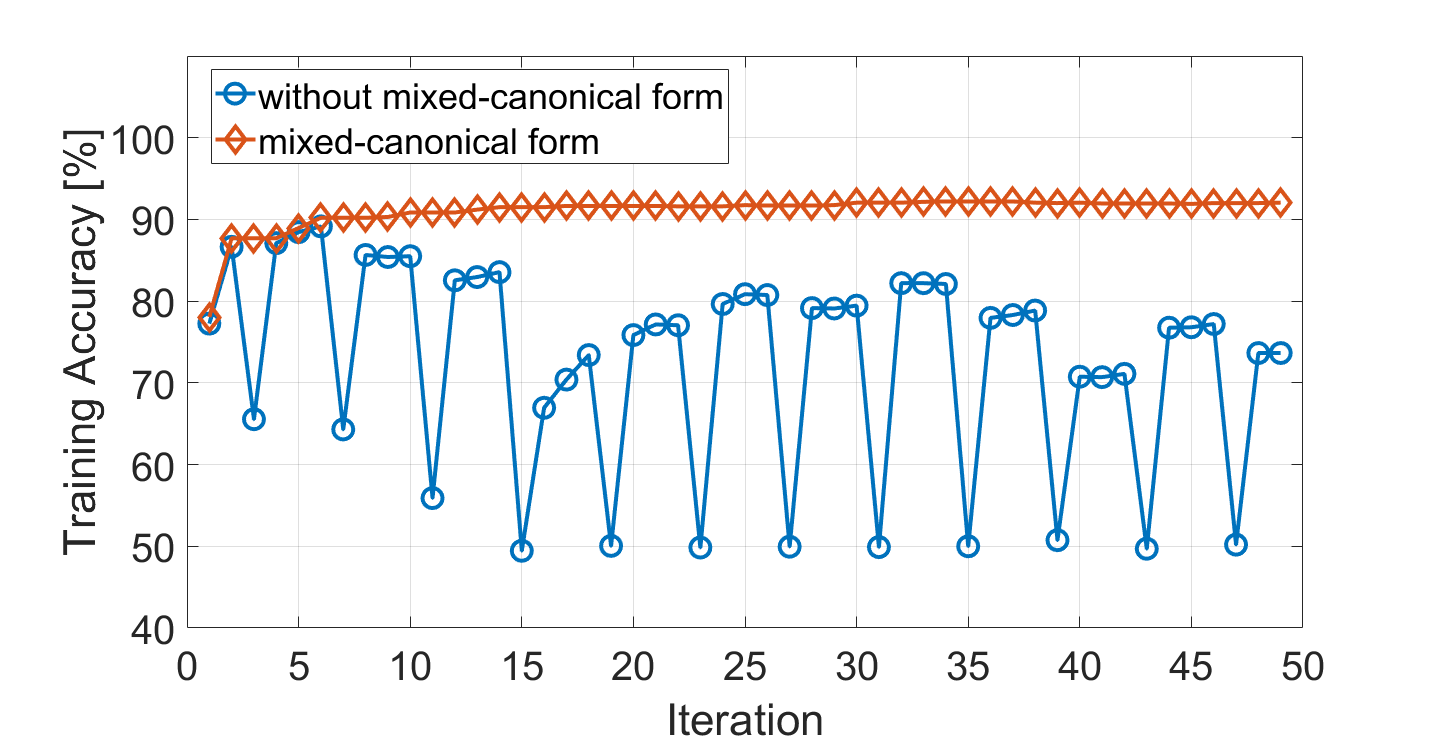}
\caption{Comparison training accuracy of STTMs trained with site-$k$-mixed-canonical form updating and without.}
\label{fig:canonical}
\end{figure}

\subsubsection{Effect of TT-Rank on Test Accuracy}
Figure~\ref{fig:experiment1_2} shows the STTM test accuracy for all tested $31$ TT-ranks when the training batch size is equal to $1000$, $1500$ and $2000$, respectively. To accommodate for the effect of the random initialization, the average test accuracy is presented over five different runs. The maximal test accuracy for these three sizes are achieved when $r_2$ is $4$, $5$, and $6$, respectively. A downward trend of all three curves can be observed for TT-ranks larger than the optimal value, indicating that higher TT-ranks may lead to overfitting. On the other hand, decreasing the TT-rank from its optimal value also decreases the test accuracy down to the STM case. An extra validation step to determine the optimal TT-ranks is therefore highly recommended. It can also be observed that the overall test accuracy improves with an increasing sample size.

\subsubsection{Updating in Site-$k$-Mixed-Canonical Form}
The effect of keeping the TT of $\ten{W}$ into site-$k$-mixed-canonical form when updating $\ten{W}^{(k)}$ is also investigated. Figure~\ref{fig:canonical} shows the training accuracy for each TT-core update iteration in Algorithm~\ref{alg:STTM}, with and without site-$k$-mixed-canonical form. Updating without the site-$k$-mixed-canonical form implies that lines $3$, $8$, $9$ and $10$ of Algorithm~\ref{alg:STTM} are not executed, which results in an oscillatory training accuracy ranging between $50\%$ and $89\%$ without any overall convergence. Updating the TT-cores $\ten{W}^{(k)}$ in a site-$k$-mixed-canonical form, however, displays a very fast convergence of the training error to around $92\%$.

\subsection{MNIST Multi-Classification}
Next, the classification accuracy of a standard SVM, STM and STTM are compared on the MNIST dataset~\cite{lecun1998gradient}, which has a training set of $60$k samples, and a testing set of $10$k samples. Each sample is a $28\times 28$ grayscale picture of a handwritten digit \{$0,\ldots,9$\}. Even though the sample structure is a 2-way tensor, we opt to reshape each sample into a $7\times 4\times 7\times 4$ tensor, as this provides us with more flexibility to choose TT-ranks when applying Algorithm~\ref{alg:STTM}. Since $10(10-1)/2=45$ binary classifiers need to be trained for this multi-classification problem, the weight vector obtained from the standard SVM is used to initialize both the STM and STTM methods. For the STM initialization, the SVM weight vector is reshaped into a $28\times 28$ matrix from which the best rank-1 approximation is used. For the STTM initialization, the SVM weight vector is reshaped into a $7\times 4\times 7\times 4$ tensor and then converted into an exact TT with ranks $r_2=r_3=5, r_4=4$ using the TT-SVD algorithm. Table~\ref{tbl1} shows the experiment setting for those three methods. All classifiers were trained for training sample batch sizes of $10$k, $20$k, $30$k and $60$k in four different experiments. The test accuracy of the different methods for different batch sizes are listed in Table~\ref{tbl2}. STTM achieves the best classification performance for all sizes. The STM again performs worse than the standard SVM due to the restrictive expressive power of the rank-1 weight matrix.
\begin{table}[tb]
\newcommand{\tabincell}[2]{\begin{tabular}{@{}#1@{}}#2\end{tabular}}
\centering
\caption{\label{mixercomparison}Experiment settings for the three methods.}
\vspace{1ex}
\begin{tabular}{@{}lrr@{}}
Method &Input Structure&\ TT-ranks  \\
\midrule
\tabincell{c}{SVM} & $784\times 1$ vector&\ NA   \\
\tabincell{c}{STM} & $28\times 28$ matrix &\ NA  \\
STTM&   $7\times 4\times 7\times 4$ tensor&\ $5,5,4$\\
\end{tabular}
\label{tbl1}
\end{table}

\begin{table}[tb]
\newcommand{\tabincell}[2]{\begin{tabular}{@{}#1@{}}#2\end{tabular}}
\centering
\caption{\label{experiment2}Test accuracy ($\%$) under different training sample sizes.}
\vspace{1ex}
\begin{tabular}{@{}lrrrr@{}}
 \multirow{2}{*}{Method} &
 \multicolumn{4}{c}{Training Sample Size} \\
   & 10k & 20k & 30k &  60k   \\
\midrule
 SVM & $91.64$ & $92.84$ & $93.28$ & $93.99$ \\
  STM & $88.36$ & $89.96$ & $89.82$ & $90.54$ \\
STTM & $\mathbf{92.27}$ & $ \mathbf{93.71}$ & $\mathbf{93.86}$ & $\mathbf{94.12}$ \\
 \end{tabular}
 \label{tbl2}
\end{table}



\section{Conclusions}
\label{sec:conclusion}
We have proposed, for the first time, a support tensor train machine (STTM) for classification. Compared with the classical support tensor machine (STM), which represents the weight parameter as a rank-1 tensor, STTM employs a more general tensor train structure to largely escalate the model expressive power. Experiments have demonstrated the superiority of STTM over standard SVM and STM in terms of classification accuracy, particularly when trained with small sample sizes. The application of kernel tricks in STTM and other numerical enhancements will be reported in a future work. 



\newpage
\bibliographystyle{named}


\end{document}